\documentclass[conference]{IEEEtran}
\IEEEoverridecommandlockouts

\usepackage{cite}
\usepackage{amsmath,amssymb}
\usepackage{graphicx}
\usepackage{booktabs}
\usepackage{xcolor}
\usepackage{enumitem}
\usepackage[hidelinks]{hyperref}

\newcommand{\rp}[2]{RP-#1/L#2}
\newcommand{\regen}[1]{Regen-#1}
\newcommand{\vine}{VINE}
\newcommand{\trustmark}{TrustMark}
\newcommand{\rivagan}{RivaGAN}
\newcommand{\dwtdctsvd}{DwtDctSvd}
\newcommand{\pt}{Paint Transformer}

\setlist[itemize]{leftmargin=1.1em,itemsep=1pt,topsep=2pt}

\begin{document}

\title{BrushOff: Stroke-Based Re-rendering as a Watermark Removal Attack on Paintings}

\author{\IEEEauthorblockN{Hao Chen}
\IEEEauthorblockA{haoc.research@gmail.com}
\and 
\IEEEauthorblockN{Junnan Xu}
\IEEEauthorblockA{junnanxu25@gmail.com}}

\maketitle

\begin{abstract}
Invisible watermarks are used to mark who owns a painting.
We show that a simple attack removes the most robust one we test.
The attack redraws the watermarked painting with brush strokes, using a pretrained stroke-based renderer.
We call this attack BrushOff.
It needs no training, no knowledge of the watermark, and no generative model.
We test four watermarks on 500 paintings from five styles.
VINE, a recent watermark built to survive image editing, survives all 14 conventional distortions we test, with at least 97.8\% detection.
Repainting lowers its detection to 1.8\%, while the output keeps a no-reference quality score close to that of the clean paintings.
Diffusion regeneration can also remove VINE, but at the same removal level it moves the image twice as far from the painting in DreamSim distance, and it changes what the painting shows.
  Because the renderer draws a countable number of strokes, we can control the attack strength step by step.
Finally, we test what kind of signal survives.
A simple probe that stores bits in the relative brightness of neighboring regions survives repainting, while a probe that uses absolute brightness does not.
The same probe fails under diffusion regeneration, while VINE fails under repainting.
Robustness to these two attacks is therefore a separate property, and benchmarks should test both.
\end{abstract}

\section{Introduction}
\label{sec:intro}

\begin{figure*}[t]
\centering
\includegraphics[width=\linewidth]{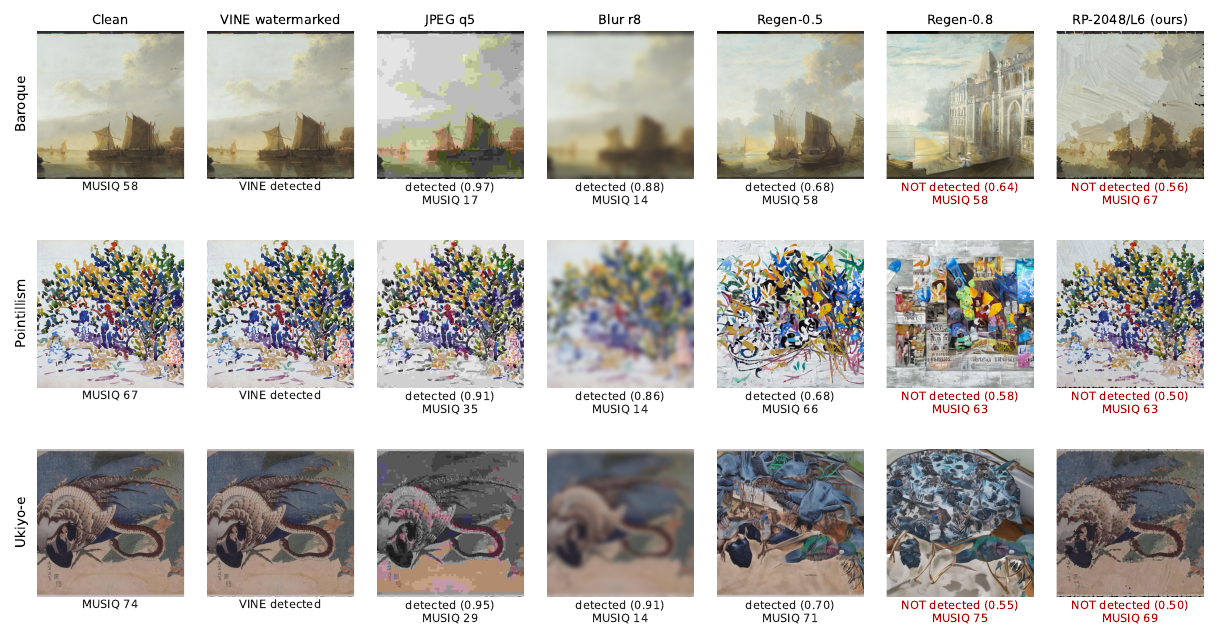}
\caption{Examples from three styles. JPEG and blur keep the \vine{} watermark but leave strong artifacts. Diffusion regeneration keeps the watermark at strength 0.5, and removes it at strength 0.8 only by changing the content. Repainting (\rp{2048}{6}) keeps the composition and removes the watermark. Below each image: whether \vine{} is detected, its bit accuracy, and MUSIQ.}
\label{fig:examples}
\end{figure*}

Invisible watermarks are a common tool to record who owns an image. Artists, museums, and platforms can embed an owner ID into a painting before sharing it online. For this to work, the mark must survive the ways people copy and modify images.

Recent watermarks are robust to many changes.
\vine{}~\cite{lu2025vine} is designed to survive not only compression and blur, but also generative editing with diffusion models.
In our tests on 500 paintings, it survives all 14 conventional distortions, including JPEG quality 5 and strong blur.

A painting can be copied in another way: it can be painted again.
A human copyist does not reproduce pixels.
They reproduce the composition, the colors, and the main shapes with brush strokes.
Stroke-based renderers~\cite{huang2019learning,liu2021paint} do the same thing automatically.
They describe an image with a set of strokes, each with a position, size, angle, and color.
We ask a simple question: does a watermark survive when a painting is redrawn with strokes?

For \vine{}, the most robust watermark we test, the answer is no.
We redraw each watermarked painting with \pt{}~\cite{liu2021paint}, a pretrained renderer, without any training. 
We call this attack BrushOff.
\vine{} drops from 100\% to 1.8\% detection.
The repainted images keep a MUSIQ quality score of 58.2, close to 59.2 for the clean paintings, and keep the composition and subject (Fig.~\ref{fig:examples}).

Repainting differs from the standard generative attack.
Diffusion regeneration~\cite{zhao2024invisible} also removes watermarks, but it relies on a model that has learned what images look like.
It can replace parts of the image with new content.
Repainting has no such prior.
Each stroke copies its color from the input.
At the same removal level, diffusion regeneration moves the image twice as far from the painting in DreamSim distance (0.559 vs.\ 0.265), and it visibly changes the content.

Repainting also gives a scale that diffusion attacks do not have.
The number of strokes is an explicit and countable bottleneck.
We can increase it step by step and see when a watermark starts to survive.

Finally, we ask what kind of signal survives repainting.
We test two simple low-frequency probes.
A probe that stores bits in the relative brightness of neighboring regions survives repainting with 83\% detection.
A probe that uses absolute brightness fails.
The surviving probe, however, fails under diffusion regeneration, while \vine{} fails under repainting.
Robustness to the two attacks is therefore a separate property.

Our contributions are:
\begin{itemize}
\item We show that stroke-based repainting removes \vine{}, a watermark that survives all conventional distortions we test, while keeping image quality close to the clean paintings.
\item We compare repainting with diffusion regeneration on 500 paintings from five styles. At the same removal level, repainting changes the painting about half as much in DreamSim distance and keeps its content.
\item We use the stroke count as a measurable bottleneck, and show with two probes that relative brightness survives repainting while absolute brightness does not. Robustness to repainting and to regeneration are separate properties.
\end{itemize}

\section{Related Work}
\label{sec:related}

\paragraph{Image watermarking}
Classical methods embed bits in a transform domain, such as DWT and DCT coefficients~\cite{cox2007digital,invisiblewatermark}.
Learned methods train an encoder and a decoder together, often with distortions in the training loop~\cite{zhu2018hidden,tancik2020stegastamp,zhang2019robust}.
\trustmark{}~\cite{bui2023trustmark} adds error correction and works at any resolution.
\vine{}~\cite{lu2025vine} uses a generative prior and is trained to survive image editing, including diffusion-based editing.
Other methods embed the watermark during generation, such as Stable Signature~\cite{fernandez2023stable}, Tree-Ring~\cite{wen2023treering}, and Gaussian Shading~\cite{yang2024gaussian}.
We focus on post-hoc watermarks, because a painting already exists before it is marked.

\paragraph{Watermark removal}
The standard removal attacks are conventional distortions and regeneration.
Zhao et al.~\cite{zhao2024invisible} show that pixel-level watermarks can be removed by encoding an image into a generative model and decoding it back.
Benchmarks such as WAVES~\cite{an2024waves} and W-Bench~\cite{lu2025vine} test watermarks against distortions, regeneration, and image editing.
Recent work uses diffusion-based image editing to remove or launder watermarks~\cite{yang2026one}.
All of these attacks rely on a learned generative prior.
Our attack does not.
This lets us separate removal by re-rendering from removal by a learned prior.

\paragraph{Protection of artworks}
Several works watermark artworks or protect artistic style~\cite{zhang2024stylemark,irwart2023,authenlora2025}.
Adversarial protections such as Glaze~\cite{shan2023glaze} can be removed by simple steps such as noisy upscaling~\cite{honig2025adversarial}.
We study a different target: invisible marks that record ownership.

\paragraph{Stroke-based rendering}
Stroke-based renderers describe an image with a sequence of brush strokes.
Learning to Paint~\cite{huang2019learning} uses reinforcement learning.
\pt{}~\cite{liu2021paint} predicts strokes in one pass with a transformer and paints from coarse to fine.
Later work predicts where to paint at each step~\cite{hu2023stroke}.
StegoNPR~\cite{lord2013stegonpr} hides data in the arrangement of non-photorealistic strokes.
That work embeds data through strokes.
We use strokes to remove data.
To our knowledge, stroke-based rendering has not been studied as a watermark removal attack.

\section{Threat Model and Repainting Attack}
\label{sec:method}

\subsection{Threat model}
\label{sec:threat}

We consider an attacker who wants to remove the provenance mark from a painting.
The attacker has one watermarked image, does not know the method, key, or payload, cannot query the detector, and does not train any model.
A typical case: a painting is shared online with an invisible owner mark, and someone downloads it, re-renders it, and posts the result as their own.

The attack succeeds if two conditions hold.
First, the detector no longer finds the watermark at a fixed false positive rate.
Second, the output still looks like the same painting, like a copy made by hand.
We therefore always report detection together with fidelity (Sec.~\ref{sec:metrics}).

\subsection{Repainting with a stroke-based renderer}
\label{sec:repaint}

Our attack redraws the watermarked image with brush strokes.
We use \pt{}~\cite{liu2021paint}, a pretrained stroke-based renderer.
We use the released model as is, without any training or tuning.

\pt{} paints from coarse to fine.
At layer $k$, it resizes the target image to $32 \cdot 2^k$ pixels per side and splits it into $32 \times 32$ patches.
For each patch, the network looks at the target and the current canvas, and predicts 8 strokes.
Each stroke has five shape parameters (center $x, y$, width, height, angle) and one RGB color.
The color is sampled from the target image at the stroke center.
A binary decision per stroke controls whether it is drawn.
For a canvas of size $S$, there are $K+1$ layers with $K = \lceil \log_2(S/32) \rceil$.
After the last layer, the official code adds one border pass.
This pass shifts the patch grid by half a patch to cover the seams between patches.

We save the canvas after every layer.
This gives a sequence of outputs from one painting run, which we call stroke levels.
Level L$\ell$ for $\ell \le K+1$ is the canvas after layer $\ell-1$.
The last level, L$(K+2)$, adds the border pass and equals the official output.
One run therefore gives all levels at no extra cost.

\paragraph{Border pass}
The border pass draws all of its strokes.
In the official code, the stroke decision in this pass is not binarized, so every predicted stroke is drawn.
We keep this behavior so that our last level matches the official output.
As a result, the last level has many more strokes than the level before it.
For this reason we use the number of strokes actually drawn, not the level index, when we study how detection depends on the attack strength (Sec.~\ref{sec:bottleneck}).

\paragraph{Canvas size}
We control stroke size by changing the canvas size.
For canvas size $S \in \{1024, 2048\}$, we upsample the $512 \times 512$ input with bicubic interpolation, paint it, and downsample the result back to $512 \times 512$ with antialiasing.
A larger canvas adds more layers, so strokes become smaller relative to the image.
The $512$, $1024$, and $2048$ canvases give 6, 7, and 8 levels.
We denote a setting by canvas size and level, for example \rp{2048}{6}.

\subsection{Why repainting removes watermarks}
\label{sec:why}

Repainting replaces every pixel with strokes.
The output is a function of the stroke parameters only.
Any signal that the strokes do not capture is lost.

\paragraph{A countable bottleneck}
The strokes form an explicit and countable bottleneck.
Each drawn stroke carries 8 numbers.
At \rp{2048}{6}, the renderer draws about 6{,}000 strokes per image, or about 48{,}000 numbers.
A $512 \times 512$ RGB image has $786{,}432$ values.
So the painting is described with about 6\% of the original values.
We can increase this budget step by step by moving to finer levels.
This lets us plot watermark survival against the number of strokes (Sec.~\ref{sec:bottleneck}).
Diffusion-based attacks do not offer such a scale, because their latent space is not easy to interpret.

\paragraph{No generative prior}
\pt{} has no generative prior over natural images.
It does not invent content.
Each stroke copies its color from the input at one location.
This separates two possible reasons for watermark removal.
Diffusion regeneration may remove a watermark because the model replaces image content with what it has learned.
Repainting can only remove a watermark by re-rendering the existing content.
If a watermark survives diffusion regeneration but not repainting, its weakness is the rendering itself, not a learned prior.

\paragraph{Cost}
The attack is cheap: about 27\,s per image at canvas size 2048 on one consumer GPU, with no training.

\section{Experimental Setup}
\label{sec:setup}

\subsection{Data}
\label{sec:data}

We use 500 paintings from WikiArt~\cite{saleh2015large}, 100 from each of five styles: Baroque, Impressionism, Pointillism, Ukiyo-e, and Color Field Painting.
We chose these styles because they differ in texture.
Baroque paintings have smooth gradients and fine detail.
Impressionism and Pointillism are made of visible brush marks and dots.
Ukiyo-e prints have flat colors and sharp outlines.
Color Field paintings have large flat regions with little texture.
We sample images at random with a fixed seed, keep those with a short side of at least 512 pixels, center-crop them to a square, and resize them to $512 \times 512$.
The WikiArt data is used for non-commercial research only.

\subsection{Watermarks}
\label{sec:watermarks}

We evaluate four post-hoc watermarks.
\vine{}~\cite{lu2025vine} is a recent learned watermark designed to survive image editing, including generative editing.
We use the released VINE-R model with a 100-bit payload.
\trustmark{}~\cite{bui2023trustmark} is a learned watermark with error correction.
We use the Q model with a 61-bit payload and the default BCH code, which gives 100 bits in total.
\rivagan{}~\cite{zhang2019robust} is a learned watermark with a 32-bit payload.
\dwtdctsvd{} is a classical frequency-domain method with a 32-bit payload.
For \rivagan{} and \dwtdctsvd{} we use the \texttt{invisible-watermark} library~\cite{invisiblewatermark}.
Every image carries the same fixed payload.

We also tested DwtDct from the same library.
It fails on 151 of the 500 paintings before any attack, and on 44 of the 100 Color Field paintings.
We report it in Table~\ref{tab:embed} but leave it out of the attack evaluation.

Table~\ref{tab:embed} reports embedding quality.
All four methods reach a mean PSNR above 36\,dB.
\vine{} has the lowest PSNR (36.9\,dB) and \trustmark{} the highest (42.4\,dB).

\begin{table}[t]
\centering
\small
\setlength{\tabcolsep}{3.5pt}
\caption{Embedding quality on 500 paintings. PSNR: mean / min. Det.: detected right after embedding. FP: detections on the 500 clean paintings.}
\label{tab:embed}
\begin{tabular}{lcccc}
\toprule
Method & Bits & PSNR (dB) & Det. & FP \\
\midrule
\vine{}      & 100     & 36.9 / 27.8 & 500 & 0 \\
\trustmark{} & 61+39   & 42.4 / 35.8 & 500 & 0 \\
\rivagan{}   & 32      & 40.5 / 39.8 & 485 & 0 \\
\dwtdctsvd{} & 32      & 37.8 / 24.7 & 500 & 0 \\
\midrule
DwtDct       & 32      & 37.8 / 24.8 & 349 & 0 \\
\bottomrule
\end{tabular}
\end{table}

\subsection{Detection}
\label{sec:detection}

We decide detection at a fixed false positive rate of $10^{-3}$.
For a payload of $n$ bits, we assume that the decoded bits of an unwatermarked image are independent and uniform.
We then pick the smallest bit accuracy $\tau$ such that the chance of reaching $\tau$ by luck is at most $10^{-3}$.
This gives $\tau = 26/32$ for \rivagan{} and \dwtdctsvd{}, and $\tau = 66/100$ for \vine{}.
For \trustmark{} we use its own decision: the watermark is detected if error correction succeeds and recovers the payload exactly.

We check the false positive rate empirically.
We run every detector on the 500 clean paintings.
None of them fires on any clean painting.

We also check each watermark right after embedding.
An image that is not detected before any attack cannot tell us anything about the attack.
We drop such images for that watermark and report them separately.
This affects only \rivagan{}, which fails on 15 of 500 paintings before any attack.
Fourteen of them are Color Field paintings, which have large flat regions and little texture to hide a signal in.

\subsection{Attacks}
\label{sec:attacks}

\paragraph{Repainting}
We run the repainting attack of Sec.~\ref{sec:repaint} with canvas sizes 512, 1024, and 2048.
This gives 21 settings, from \rp{512}{1} to \rp{2048}{8}.
For \vine{}, we run all settings on all 500 paintings.
Canvas size 2048 is slow (Sec.~\ref{sec:why}).
For the other three watermarks, we therefore run it on a subset of 100 paintings, the first 20 of each style.

\paragraph{Conventional distortions}
We apply 14 standard distortions:
JPEG compression with quality 90, 70, 50, 30, 10, and 5;
Gaussian blur with radius 1, 2, 4, 6, and 8 pixels;
and down-scaling by a factor of 0.5, 0.25, and 0.125 followed by up-scaling back to $512 \times 512$ with bicubic interpolation.
We save all outputs as PNG, so no further compression is added.

\paragraph{Diffusion regeneration}
We compare against diffusion regeneration~\cite{zhao2024invisible}, the standard generative removal attack.
The image is encoded with the Stable Diffusion VAE, noised to a given strength, and denoised back.
We use Stable Diffusion 2.1-base~\cite{rombach2022ldm} in image-to-image mode with an empty prompt, no classifier-free guidance, 50 denoising steps, and a fixed seed.
We test strengths 0, 0.1, 0.2, 0.3, and 0.5 on all four watermarks.
Strength 0 is a VAE round trip without noise.
For \vine{}, we add strengths 0.6, 0.7, and 0.8, because \vine{} still survives at 0.5.
We denote a setting by its strength, for example \regen{0.5}.

\subsection{Fidelity metrics}
\label{sec:metrics}

We measure how close the attacked image stays to the clean painting.
All metrics compare against the clean painting, not the watermarked one.
We use three metrics, because no single one captures what we need.

DISTS~\cite{ding2020dists} measures structure and texture similarity.
DreamSim~\cite{fu2023dreamsim} measures perceptual similarity at a higher level, closer to whether two images show the same thing.
For both, lower means closer to the clean painting.
MUSIQ~\cite{ke2021musiq} is a no-reference quality score.
It measures whether an image looks free of artifacts such as blur, noise, and block patterns.
Higher is better.
The clean paintings have a mean MUSIQ of 59.2.
We also report PSNR for reference.

MUSIQ does not check whether the content matches the painting: a coarse repainting with a few hundred strokes can score high.
We therefore always read MUSIQ together with DreamSim.

In a pilot study, fidelity barely depended on which watermark the image carried.
We therefore compute fidelity on the \vine{} images and use it for all methods.

\paragraph{Statistics}
We report 95\% confidence intervals for detection rates.
We compute them with a bootstrap over images, using 2{,}000 resamples and the percentile method.

\section{Results}
\label{sec:results}

Table~\ref{tab:main} gives the main numbers.

\begin{table*}[t]
\centering
\small
\setlength{\tabcolsep}{4.5pt}
\caption{Fidelity and detection under each attack. Fidelity is measured against the clean painting. The clean paintings have a mean MUSIQ of 59.2. Detection rates use a fixed false positive rate of $10^{-3}$. $^\dagger$: 100-image subset. --: not run.}
\label{tab:main}
\begin{tabular}{l r cccc cccc}
\toprule
 & & \multicolumn{4}{c}{Fidelity} & \multicolumn{4}{c}{Detection rate} \\
\cmidrule(lr){3-6}\cmidrule(lr){7-10}
Attack & Strokes & DISTS$\downarrow$ & DreamSim$\downarrow$ & MUSIQ$\uparrow$ & PSNR$\uparrow$ & \vine{} & \trustmark{} & \rivagan{} & \dwtdctsvd{} \\
\midrule
JPEG q10          & -- & 0.267 & 0.111 & 42.7 & 25.5 & 1.000 & 0.068 & 0.082 & 0.000 \\
JPEG q5           & -- & 0.339 & 0.230 & 32.3 & 23.1 & 0.992 & 0.000 & 0.002 & 0.000 \\
Blur r4           & -- & 0.353 & 0.168 & 16.7 & 23.4 & 1.000 & 0.954 & 0.705 & 0.562 \\
Blur r8           & -- & 0.470 & 0.355 & 15.1 & 21.5 & 1.000 & 0.050 & 0.000 & 0.058 \\
Resize 0.125      & -- & 0.337 & 0.163 & 17.9 & 23.7 & 0.978 & 0.826 & 0.695 & 0.890 \\
\midrule
\regen{0.1}       & -- & 0.124 & 0.061 & 57.7 & 23.0 & 0.998 & 0.000 & 0.002 & 0.092 \\
\regen{0.3}       & -- & 0.168 & 0.122 & 57.9 & 20.6 & 0.950 & 0.000 & 0.000 & 0.012 \\
\regen{0.5}       & -- & 0.221 & 0.256 & 59.1 & 18.3 & 0.474 & 0.000 & 0.000 & 0.000 \\
\regen{0.8}       & -- & 0.325 & 0.559 & 63.2 & 14.4 & 0.018 & --    & --    & --    \\
\midrule
\rp{512}{6}       & 4{,}249  & 0.383 & 0.280 & 59.5 & 20.1 & 0.532 & 0.000 & 0.126 & 0.262 \\
\rp{1024}{6}      & 6{,}045  & 0.349 & 0.267 & 57.9 & 19.9 & 0.030 & 0.000 & 0.196 & 0.158 \\
\rp{2048}{6}      & 6{,}042  & 0.349 & 0.265 & 58.2 & 19.9 & \textbf{0.018} & 0.000$^\dagger$ & 0.226$^\dagger$ & 0.180$^\dagger$ \\
\rp{2048}{7}      & 16{,}392 & 0.309 & 0.182 & 58.3 & 21.6 & 0.618 & 0.000$^\dagger$ & 0.656$^\dagger$ & 0.290$^\dagger$ \\
\rp{2048}{8}      & 50{,}192 & 0.258 & 0.116 & 57.1 & 23.6 & 1.000 & 0.030$^\dagger$ & 0.849$^\dagger$ & 0.470$^\dagger$ \\
\bottomrule
\end{tabular}
\end{table*}

\subsection{Repainting removes VINE}
\label{sec:res-vine}

\vine{} survives every conventional distortion we test.
Its detection rate stays at or above 97.8\% under all 14 distortions, including JPEG quality 5 and blur radius 8 (Table~\ref{tab:main}).
These two distortions are already strong.
JPEG q5 drops MUSIQ from 59.2 to 32.3, and blur radius 8 drops it to 15.1.

Repainting removes \vine{}.
At \rp{2048}{6}, the detection rate drops to 1.8\% (95\% CI: 0.8--3.2\%).
At \rp{1024}{6}, it drops to 3.0\%.

\begin{figure*}[t]
\centering
\includegraphics[width=\linewidth]{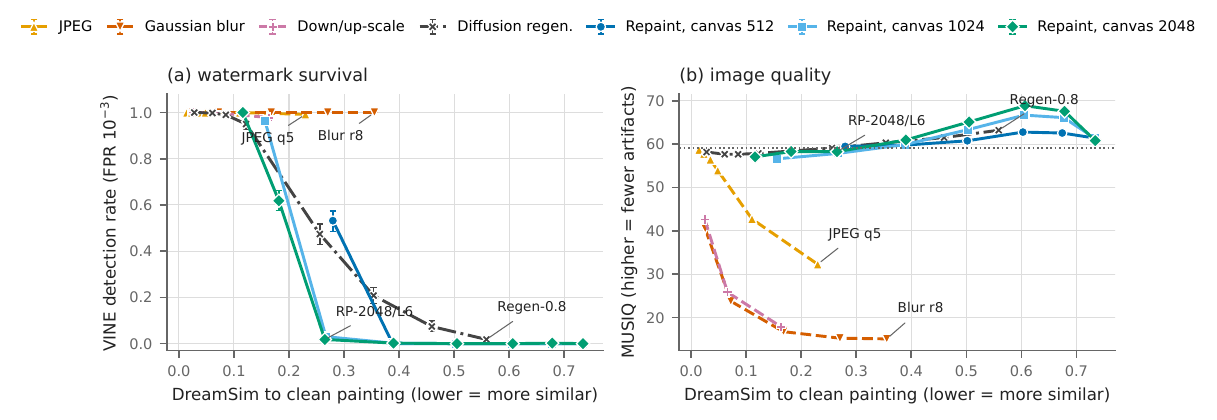}
\caption{(a) \vine{} detection rate and (b) MUSIQ against DreamSim distance to the clean painting. Each curve is one attack family at increasing strength. Error bars: 95\% bootstrap confidence intervals. Dotted line in (b): mean MUSIQ of the clean paintings.}
\label{fig:fidelity}
\end{figure*}

The repainted images still look like normal paintings.
\rp{2048}{6} keeps a MUSIQ of 58.2, close to the 59.2 of the clean paintings.
Its DreamSim to the clean painting is 0.265.
This is close to JPEG q5 (0.230) and lower than blur radius 8 (0.355).
In other words, at a similar perceptual distance, JPEG and blur leave visible artifacts and keep the watermark, while repainting gives a clean-looking image and removes the watermark.
Fig.~\ref{fig:fidelity} shows this for all settings.
Fig.~\ref{fig:examples} shows examples.

Coarser levels remove all watermarks but change the painting too much.
At levels L1 to L5, no watermark is detected in more than 6\% of images at any canvas size.
But DreamSim is above 0.35 at these levels, and the images lose much of their detail.
We therefore focus on L6 and finer levels.

\subsection{Comparison with diffusion regeneration}
\label{sec:res-regen}

Diffusion regeneration can also remove \vine{}, but it needs to change the image much more.
At strength 0.5, \vine{} is still detected in 47.4\% of images (95\% CI: 42.8--51.8\%).
Detection only falls to 1.8\%, the level reached by \rp{2048}{6}, at strength 0.8.

At the same removal level, repainting stays closer to the painting.
\regen{0.8} has a DreamSim of 0.559, while \rp{2048}{6} has 0.265.
Its PSNR is 14.4\,dB, while \rp{2048}{6} has 19.9\,dB.
DISTS is similar for the two (0.325 and 0.349).

At a similar DreamSim, repainting removes far more.
\regen{0.5} and \rp{2048}{6} have almost the same DreamSim (0.256 and 0.265) and MUSIQ (59.1 and 58.2).
\vine{} is detected in 47.4\% of images after \regen{0.5}, but in only 1.8\% after \rp{2048}{6}.

Diffusion regeneration also changes what the painting shows.
In Fig.~\ref{fig:examples}, \regen{0.5} turns the dragon in the Ukiyo-e print and the bush in the Pointillist painting into different shapes, while \vine{} is still detected.
Repainting keeps the composition and the subject, while \vine{} is gone.

MUSIQ does not penalize this change.
The MUSIQ of diffusion regeneration rises with strength, from 58.2 at strength 0 to 63.2 at 0.8.
The model adds sharp details that were not in the painting, and MUSIQ rewards them.
This is why we read MUSIQ together with DreamSim.

We see a simple reason for this result.
\vine{} is trained to survive generative editing~\cite{lu2025vine}, so diffusion regeneration is close to what it was built for.
Repainting has no generative prior and renders the image in a different way.
\vine{} was not trained for this kind of change.

\subsection{Detection and the number of strokes}
\label{sec:bottleneck}

\begin{figure*}[t]
\centering
\includegraphics[width=\linewidth]{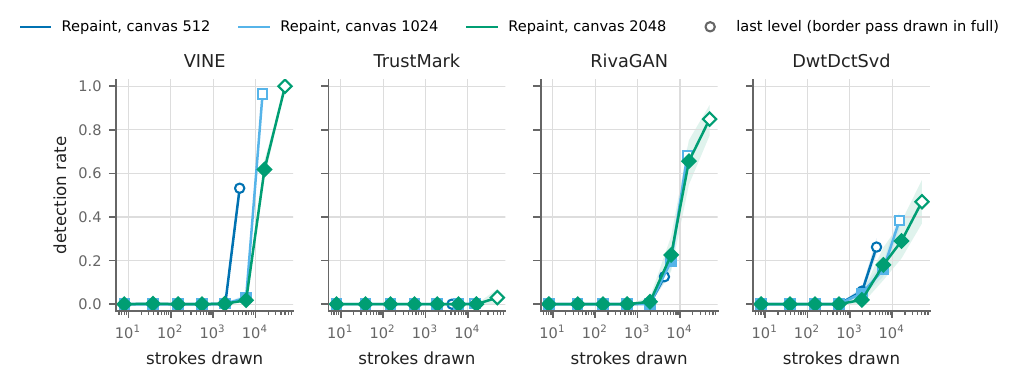}
\caption{Detection rate against the number of strokes drawn, for each watermark and canvas size. Hollow markers: last level, where the border pass draws all of its strokes. Shaded bands: 95\% bootstrap confidence intervals. \trustmark{}, \rivagan{}, and \dwtdctsvd{} at canvas size 2048 use a 100-image subset.}
\label{fig:strokes}
\end{figure*}

For the three weaker watermarks, detection follows the number of strokes.
Fig.~\ref{fig:strokes} plots detection against the number of strokes drawn.
For \rivagan{} and \dwtdctsvd{}, the curves from the three canvas sizes almost overlap.
Detection rises only after a few thousand strokes.
\trustmark{} stays near zero at every level.

\vine{} shows a sharp transition.
It is detected in at most 0.2\% of images at about 1{,}900 strokes or fewer (L5 at every canvas size).
It is detected in 62\% of images at about 16{,}000 strokes (\rp{2048}{7}), and in all images at about 50{,}000 strokes (\rp{2048}{8}). So more strokes is not always better for the attacker: the best setting is the finest level that still removes the watermark, here \rp{2048}{6}.

For \vine{}, the stroke count is not the whole story.
\rp{512}{6} draws about 4{,}200 strokes, fewer than \rp{2048}{6} with about 6{,}000.
Yet \vine{} is detected in 53\% of images after \rp{512}{6}, and in only 1.8\% after \rp{2048}{6}.
Likewise, \rp{1024}{7} and \rp{2048}{7} draw a similar number of strokes (14{,}800 and 16{,}400), but give 96\% and 62\%.
In both pairs, the setting with the larger canvas, and therefore smaller strokes, removes more.
Both \rp{512}{6} and \rp{1024}{7} are last levels, where the border pass draws all of its strokes (Sec.~\ref{sec:repaint}).
So the size and placement of strokes also matter, not only their number.

\subsection{Differences between styles}
\label{sec:res-style}

At \rp{2048}{6}, repainting removes \vine{} in every style.
The detection rate is at most 7\% in each of the five styles.

At the next level, styles differ.
At \rp{2048}{7}, \vine{} is detected in 40\% of Color Field paintings and in 87\% of Pointillist paintings.
The other three styles fall between 55\% and 65\%.

Diffusion regeneration shows a different pattern.
At \regen{0.5}, \vine{} survives best on Color Field (70\%) and Baroque (67\%), and worst on Ukiyo-e (30\%) and Impressionism (29\%).

\subsection{The weaker watermarks}
\label{sec:res-weak}

Repainting changes the robustness ranking of the weaker watermarks.
\trustmark{} survives blur radius 4 (95\%) and down-scaling to 0.125 (83\%), but every repainting setting removes it.
Even at \rp{2048}{8}, its detection rate is 3\%.
\rivagan{} behaves the other way.
It fails under blur radius 8 (0\%), but survives the finest repainting (85\% at \rp{2048}{8}).
Robustness to conventional distortions therefore does not predict robustness to repainting.

For these three watermarks, repainting is not the cheapest attack.
JPEG q10 removes all three at a DreamSim of 0.111.
\regen{0.1} lowers all three to at most 9\% at a DreamSim of 0.061.
Even a VAE round trip without noise (\regen{0}) lowers \rivagan{} to 32\%.
For RivaGAN and DwtDctSvd, repainting reaches similar removal only at the coarse levels, where DreamSim is above 0.35.
\trustmark{} is the exception: every repainting setting removes it, down to a DreamSim of 0.116.
The advantage of repainting is specific to \vine{}, the watermark built to resist generative editing.

\section{What Survives Repainting?}
\label{sec:probe}

Repainting keeps the composition of a painting: where things are, and which areas are lighter or darker.
If a watermark is carried by this coarse structure, it may survive.
We test this idea with two simple low-frequency probes.
The probes are not new watermarking methods.
They are textbook schemes that we use only to test the idea.

\subsection{Two probes}
\label{sec:probe-design}

Both probes split the image into a $G \times G$ grid of blocks and change only the mean brightness of each block.
The change is added through a smooth Hann window inside each block, so no block edges appear.
The same change is added to R, G, and B, so only brightness changes.
Both probes carry the same 32-bit payload, repeated over the blocks with a keyed permutation.
We use the same detection threshold as for the other 32-bit methods ($26/32$).

\paragraph{QIM: absolute brightness}
The first probe stores one bit in the absolute mean brightness of each block.
It uses quantization index modulation~\cite{chen2001qim} with step $\Delta$.
We test $G \in \{8, 16\}$ and $\Delta \in \{8, 16\}$.

\paragraph{Pair: relative brightness}
The second probe stores one bit in the brightness difference between two neighboring blocks.
For bit 1, the left block must be brighter than the right block by at least $T$; for bit 0, darker by at least $T$.
Pairs that already satisfy this are left unchanged.
We cap the change of any block at $2T$, so very large natural differences are not forced.
We use $G = 16$, which gives 128 pairs and 4 repetitions per bit, and test $T \in \{4, 8\}$.
At decoding, each pair gives one vote, $\mathrm{clip}(d/T, -1, 1)$, where $d$ is the measured difference.

We changed this decoder once.
Our first version summed the raw differences, and large natural differences hid the embedded votes: only 36 of 100 paintings were detected right after embedding with $T = 8$.
The clipped vote fixes this without changing the false positive rate, and we use it throughout.

\subsection{Setup}

We embed both probes in 100 paintings, the first 20 of each style.
We attack them with the same repainting, conventional, and regeneration settings as before.
We count an image as missed if the probe fails right after embedding.
So the detection rates below are over all 100 images.

Both probes have no false positives in practice.
On the 500 clean paintings, five of the six configurations never fire.
QIM with $G = 16, \Delta = 8$ fires once, which is in line with the designed rate of $10^{-3}$.

The probes trade visibility for robustness.
QIM with $\Delta = 8$ and pair with $T = 4$ reach a PSNR of about 37.5\,dB, close to \vine{} (36.9\,dB).
QIM with $\Delta = 16$ and pair with $T = 8$ reach only 31.5\,dB and 30.6\,dB.
Pair with $T = 4$ cannot be written into every painting: it is detected right after embedding in 60 of 100 paintings.
With $T = 8$ this rises to 96 of 100.

\subsection{Results}

Relative brightness survives repainting.
Pair with $T = 8$ is detected in 83\% of images after \rp{2048}{6} (95\% CI: 76--89\%), where \vine{} is detected in 1.8\%.
It is still detected in 77\% of images at \rp{512}{5}, which draws only about 1{,}900 strokes.
It also survives all conventional distortions, with at least 91\% detection.

Absolute brightness does not survive repainting.
All four QIM configurations are detected in at most 5\% of images at L6 and finer levels.
They are robust to conventional distortions, with 66--98\% detection under blur radius 8 and 71--97\% under JPEG q5.
Their bit accuracy after repainting is often below chance.
For example, QIM with $G = 8, \Delta = 16$ has a bit accuracy of 0.31 after \rp{512}{6}.
A bit accuracy far below 0.5 means that many bits flip in the same direction.
So repainting does not erase the block brightness.
It shifts it, and absolute codes cannot tolerate this shift.
The pair probe compares two neighboring blocks, so a shared shift cancels out.

The probe also holds at a matched PSNR.
Pair with $T = 4$ has about the same PSNR as \vine{}.
It is detected in 29\% of images after \rp{2048}{6} (95\% CI: 22--36\%), compared with 1.8\% for \vine{}.

\begin{figure}[t]
\centering
\includegraphics[width=\linewidth]{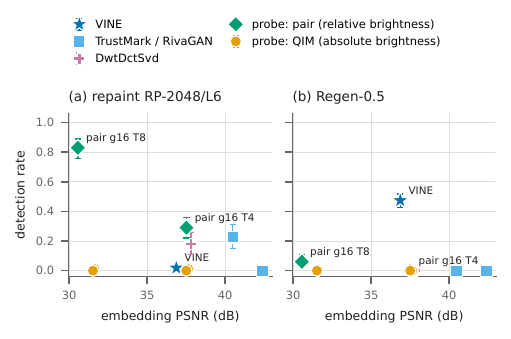}
\caption{Detection rate against embedding PSNR after (a) \rp{2048}{6} and (b) \regen{0.5}. Probes use 100 paintings, and embedding failures count as misses. Error bars: 95\% bootstrap CIs.}
\label{fig:probe}
\end{figure}

Robustness to repainting and to diffusion regeneration are separate properties.
The pair probe that survives repainting fails under diffusion regeneration: 6\% after \regen{0.5}.
\vine{} shows the opposite pattern: 1.8\% after \rp{2048}{6}, but 47\% after \regen{0.5}.
Fig.~\ref{fig:probe} shows this for all methods.
No method we test survives both.

\section{Discussion and Limitations}
\label{sec:discussion}

\paragraph{Robustness has more than one axis}
Our results show that a watermark can be robust to one kind of re-rendering and fragile to another.
\vine{} resists diffusion regeneration but not repainting.
Our pair probe resists repainting but not diffusion regeneration.
Robustness to conventional distortions does not predict either one: \trustmark{} survives blur but not repainting, and \rivagan{} the other way round.
Benchmarks that test only conventional distortions and generative regeneration~\cite{lu2025vine,an2024waves} therefore miss a real attack.
We suggest adding non-generative re-rendering, such as stroke-based repainting, to these benchmarks.
A claim that a watermark is hard to remove should also be tested against repainting.

\paragraph{Where a robust signal could live}
Our probe suggests that a signal carried by relative brightness between regions survives repainting.
This is the level that strokes reproduce.
The probe pays for this with visibility: with $T = 8$ it is about 6\,dB lower in PSNR than \vine{}.
It also fails under diffusion regeneration.
Combining such a signal with a learned watermark may cover both attacks.
We leave this to future work.

\paragraph{Limitations}
Our study has several limits.
\begin{itemize}
\item We test one stroke-based renderer, \pt{}. Other renderers, such as Learning to Paint~\cite{huang2019learning}, may behave differently.
\item Repainting changes the texture of a painting. The output has visible brush strokes, so it is a convincing copy, not an exact one. For uses that need an exact copy, this attack is less useful.
\item The repainted images have small dark spots along the right and bottom borders. They come from the black initial canvas of the official code, where no stroke covers it. They do not affect our conclusions, but a stronger attacker could easily remove them.
\item Canvas size 2048 for the three weaker watermarks uses 100 images, and fidelity is computed on the \vine{} images only.
\item We changed the pair decoder once after seeing the first results (Sec.~\ref{sec:probe-design}).
\item We do not test watermarks embedded during image generation, such as Tree-Ring~\cite{wen2023treering} or Gaussian Shading~\cite{yang2024gaussian}.
\item We only study paintings. On photographs, the repainted output would look like a painting, so our threat model does not apply.
\end{itemize}

\paragraph{Ethics}
This paper describes a way to remove watermarks from paintings.
The attack uses only a public pretrained model, and anyone can run it already.
We think it is better for watermark designers to know about it.
We follow the licenses of the data and models: WikiArt and the \vine{} weights are for non-commercial research only.

\section{Conclusion}
\label{sec:conclusion}

Repainting a painting with brush strokes removes \vine{}, which survives all conventional distortions we tested.
The attack needs no training and no generative model, and the output still looks like the painting.
At the same removal level, diffusion regeneration changes the painting about twice as much.
No watermark we tested resists both attacks, so we hope future benchmarks include stroke-based repainting.

\bibliographystyle{IEEEtran}
\bibliography{refs}

\end{document}